# Variational Relevance Vector Machines

Christopher M. Bishop           Michael E. Tipping

Microsoft Research
St. George House, 1 Guildhall Street
Cambridge CB2 3NH, U.K.
{cmbishop,mtipping}@microsoft.com

## Abstract

The Support Vector Machine (SVM) of Vapnik [9] has become widely established as one of the leading approaches to pattern recognition and machine learning. It expresses predictions in terms of a linear combination of kernel functions centred on a subset of the training data, known as support vectors.

Despite its widespread success, the SVM suffers from some important limitations, one of the most significant being that it makes point predictions rather than generating predictive distributions. Recently Tipping [8] has formulated the Relevance Vector Machine (RVM), a probabilistic model whose functional form is equivalent to the SVM. It achieves comparable recognition accuracy to the SVM, yet provides a full predictive distribution, and also requires substantially fewer kernel functions.

The original treatment of the RVM relied on the use of type II maximum likelihood (the 'evidence framework') to provide point estimates of the hyperparameters which govern model sparsity. In this paper we show how the RVM can be formulated and solved within a completely Bayesian paradigm through the use of variational inference, thereby giving a posterior distribution over both parameters and hyperparameters. We demonstrate the practicality and performance of the variational RVM using both synthetic and real world examples.

## 1 RELEVANCE VECTORS

Many problems in machine learning fall under the heading of supervized learning, in which we are given a set of input vectors $X = \{\mathbf{x}_n\}_{n=1}^N$ together with corresponding target values $T = \{t_n\}_{n=1}^N$. The goal is to use this training data, together with any pertinent prior knowledge, to make predictions of $t$ for new values of $\mathbf{x}$. We can distinguish two distinct cases: *regression*, in which $t$ is a continuous variable, and *classification*, in which $t$ belongs to a discrete set.

Here we consider models in which the prediction $y(\mathbf{x}, \mathbf{w})$ is expressed as a linear combination of basis functions $\phi_m(\mathbf{x})$ of the form

$$y(\mathbf{x}, \mathbf{w}) = \sum_{m=0}^M w_m \phi_m(\mathbf{x}) = \mathbf{w}^{\mathrm{T}} \boldsymbol{\phi} \qquad (1)$$

where the $\{w_m\}$ are the parameters of the model, and are generally called *weights*.

One of the most popular approaches to machine learning to emerge in recent years is the Support Vector Machine (SVM) of Vapnik [9]. The SVM uses a particular specialization of (1) in which the basis functions take the form of *kernel* functions, one for each data point $\mathbf{x}_m$ in the training set, so that $\phi_m(\mathbf{x}) = K(\mathbf{x}, \mathbf{x}_m)$, where $K(\cdot, \cdot)$ is the kernel function. The framework which we develop in this paper is much more general and applies to any model of the form (1). However, in order to facilitate direct comparisons with the SVM, we focus primarily on the use of kernels as the basis functions.

Point estimates for the weights are determined in the SVM by optimization of a criterion which simultaneously attempts to fit the training data while at the same time minimizing the 'complexity' of the function $y(\mathbf{x}, \mathbf{w})$. The result is that some proportion of the weights are set to zero, leading to a sparse model in which predictions, governed by (1), depend only on a subset of the kernel functions.

The SVM framework is found to yield good predictive performance for a broad range of practical applications, and is widely regarded as the state of the art in pattern recognition. However, the SVM suffers from some important drawbacks. Perhaps the most significant of these is that it is a non-Bayesian approach which makes explicit classifications (or point predictions in the case of regression) for new inputs. As is well known, there are numerous advantages to predicting the posterior probability of class membership (or a predictive conditional distribution in the case of regression). These include the optimal compensation for skewed loss matrices or unequal class distributions, the opportunity to improve performance by rejection of the more ambiguous examples, and the fusion of outputs with other probabilistic sources information before applying decision criteria.

Recently Tipping [8] introduced the *Relevance Vector Machine* (RVM) which makes probabilistic predictions and yet which retains the excellent predictive performance of the support vector machine. It also preserves the sparseness property of the SVM. Indeed, for a wide variety of test problems it actually leads to models which are dramatically sparser than the corresponding SVM, while sacrificing little if anything in the accuracy of prediction.

For regression problems, the RVM models the conditional distribution of the target variable, given an input vector $\mathbf{x}$, as a Gaussian distribution of the form

$$P(t|\mathbf{x}, \mathbf{w}, \tau) = \mathcal{N}(t|y(\mathbf{x}, \mathbf{w}), \tau^{-1}) \qquad (2)$$

where we use $\mathcal{N}(\mathbf{z}|\mathbf{m}, \mathbf{S})$ to denote a multi-variate Gaussian distribution over $\mathbf{z}$ with mean $\mathbf{m}$ and covariance $\mathbf{S}$. In (2) $\tau$ is the inverse 'noise' parameter, and the conditional mean $y(\mathbf{x}, \mathbf{w})$ is given by (1). Assuming an independent, identically distributed data set $X = \{\mathbf{x}_n\}$, $T = \{t_n\}$ the likelihood function can be written

$$P(T|X, \mathbf{w}, \tau) = \prod_{n=1}^{N} P(t_n|\mathbf{x}_n, \mathbf{w}, \tau). \qquad (3)$$

The parameters $\mathbf{w}$ are given a Gaussian prior

$$P(\mathbf{w}|\boldsymbol{\alpha}) = \prod_{m=0}^{N} \mathcal{N}(w_m|0, \alpha_m^{-1}) \qquad (4)$$

where $\boldsymbol{\alpha} = \{\alpha_m\}$ is a vector of hyperparameters, with one hyperparameter $\alpha_m$ assigned to each model parameter $w_m$. In the original RVM of Tipping [8] values for these hyperparameters are estimated using the framework of type-II maximum likelihood [1] in which the marginal likelihood $P(T|X, \boldsymbol{\alpha}, \tau)$ is maximized with respect to $\boldsymbol{\alpha}$ and $\tau$. Evaluation of this marginal likelihood requires integration over the model parameters

$$P(T|X, \boldsymbol{\alpha}, \tau) = \int P(T|X, \mathbf{w}, \tau) P(\mathbf{w}|\boldsymbol{\alpha}) \, d\mathbf{w}. \qquad (5)$$

Since this involves the convolution of two exponential-quadratic functions the integration can be performed analytically, giving

$$P(T|X, \boldsymbol{\alpha}, \tau) = \mathcal{N}(\mathbf{t}|\mathbf{0}, \mathbf{S}) \qquad (6)$$

where $\mathbf{t} = (t_1, \ldots, t_N)$ and

$$\mathbf{S} = \tau^{-1}\mathbf{I} + \boldsymbol{\Phi}\mathbf{A}^{-1}\boldsymbol{\Phi}^{\mathrm{T}} \qquad (7)$$

in which $\mathbf{I}$ is the $N \times N$ unit matrix, $\mathbf{A} = \text{diag}(\alpha_m)$, and $\boldsymbol{\Phi}$ is the $N \times (N+1)$ *design matrix* with columns $\phi_m$, so that $(\boldsymbol{\Phi})_{nm} = \phi(\mathbf{x}_n; \mathbf{x}_m)$. Maximization of (6) with respect to the $\{\alpha_m\}$ can be performed efficiently using an iterative re-estimation procedure obtained by setting the derivatives of the marginal log likelihood to zero. During the process of this optimization many of the $\alpha_m$ are driven to large values, so that the corresponding model parameters $w_m$ are effectively pruned out. The corresponding terms can be omitted from the trained model represented by (1), with the training data vectors $\mathbf{x}_n$ associated with the remaining kernel functions being termed 'relevance vectors'. Insight into this pruning process is given in Section 3. A similar re-estimation procedure is used to optimize $\tau$ simultaneously with the $\alpha_m$ parameters.

In the classification version of the relevance vector machine the conditional distribution of targets is given by

$$P(t|\mathbf{x}, \mathbf{w}) = \sigma(y)^t [1 - \sigma(y)]^{1-t} \qquad (8)$$

where $\sigma(y) = (1 + \exp(-y))^{-1}$ and $y(\mathbf{x}, \mathbf{w})$ is given by (1). Here we confine attention to the case $t \in \{0, 1\}$. Assuming independent, identically distributed data, we obtain the likelihood function in the form

$$P(T|X, \mathbf{w}) = \prod_{n=1}^{N} \sigma(y_n)^{t_n} [1 - \sigma(y_n)]^{1-t_n}. \qquad (9)$$

As before, the prior over the weights takes the form (4). However, the integration required by (5) in order to evaluate the marginal likelihood can no longer be performed analytically. Tipping [8] therefore used a local Gaussian approximation to the posterior distribution of the weights. Optimization of the hyperparameters can then be performed using a re-estimation framework, alternating with re-evaluation of the mode of the posterior, until convergence.

As we have seen, the standard relevance vector machine of Tipping [8] estimates point values for the hyperparameters. In this paper we seek a more complete Bayesian treatment of the RVM through exploitation of variational methods.

## 2 VARIATIONAL INFERENCE

In a general probabilistic model we can partition the stochastic variables into those corresponding to the observed data, denoted $D$, and the remaining unobserved variables denoted $\boldsymbol{\theta}$. The marginal probability of the observed data (the model 'evidence') is obtained by integrating over $\boldsymbol{\theta}$

$$P(D) = \int P(D, \boldsymbol{\theta}) \, d\boldsymbol{\theta}. \qquad (10)$$

This integration will, for almost any non-trivial model, be analytically intractable. Variational methods [4] address this problem by introducing a distribution $Q(\boldsymbol{\theta})$, which (for arbitrary choice of $Q$) allows the marginal log likelihood to be decomposed into two terms [6]

$$\ln P(D) = \mathcal{L}(Q) + \mathrm{KL}(Q\|P) \qquad (11)$$

where

$$\mathcal{L} = \int Q(\boldsymbol{\theta}) \ln \frac{P(D, \boldsymbol{\theta})}{Q(\boldsymbol{\theta})} \, d\boldsymbol{\theta} \qquad (12)$$

and $\mathrm{KL}(Q\|P)$ is the Kullback-Leibler divergence between $Q(\boldsymbol{\theta})$ and the posterior distribution $P(\boldsymbol{\theta}|D)$, and is given by

$$\mathrm{KL}(Q\|P) = -\int Q(\boldsymbol{\theta}) \ln \frac{P(\boldsymbol{\theta}|D)}{Q(\boldsymbol{\theta})} \, d\boldsymbol{\theta}. \qquad (13)$$

Since $\mathrm{KL}(Q\|P) \geq 0$, it follows that $\mathcal{L}(Q)$ is a rigorous lower bound on $\ln P(D)$. Furthermore, since the left hand side of (11) is independent of $Q$, maximizing $\mathcal{L}(Q)$ is equivalent to minimizing $\mathrm{KL}(Q\|P)$, and therefore $Q(\boldsymbol{\theta})$ represents an approximation to the posterior distribution $P(\boldsymbol{\theta}|D)$.

The significance of this transformation is that, for a suitable choice for the $Q$ distribution, the quantity $\mathcal{L}(Q)$ may be tractable to compute, even though the original model evidence function is not. The goal in a variational approach is therefore to choose a suitable form for $Q(\boldsymbol{\theta})$ which is sufficiently simple that the lower bound $\mathcal{L}(Q)$ can readily be evaluated and yet which is sufficiently flexible that the bound is reasonably tight. In practice we choose some family of $Q$ distributions and then seek the best approximation within this family by maximizing the lower bound with respect to $Q$. One approach would be to assume some specific parameterized functional form for $Q$ and then to optimize $\mathcal{L}$ with respect to the parameters of the distribution. Here we adopt an alternative procedure, following [10], and consider a factorized form over the component variables $\{\theta_i\}$ in $\boldsymbol{\theta}$, so that

$$Q(\boldsymbol{\theta}) = \prod_i Q_i(\theta_i). \qquad (14)$$

The lower bound can then be maximized over all possible factorial distributions by performing a *free-form* maximization over the $Q_i$, leading to the following result

$$Q_i(\theta_i) = \frac{\exp \langle \ln P(D, \boldsymbol{\theta}) \rangle_{k \neq i}}{\int \exp \langle \ln P(D, \boldsymbol{\theta}) \rangle_{k \neq i} \, d\theta_i} \qquad (15)$$

where $\langle \cdot \rangle_{k \neq i}$ denotes an expectation with respect to the distributions $Q_k(\theta_k)$ for all $k \neq i$. It is easily shown that, if the probabilistic model is expressed as a directed acyclic graph with a node for each of the factors $Q_i(\theta_i)$, then the solution for $Q_i(\theta_i)$ depends only on the $Q$ distributions for variables which are in the Markov blanket of the node $i$ in the graph.

Note that (15) represents an implicit solution for the factors $Q_i(\theta_i)$ since the right hand side depends on moments with respect to the $Q_{k \neq i}$. For conjugate conditional distributions (e.g. linear-Gaussian models with Gamma priors, in the case of continuous variables) this leads to standard distributions for which the required moments are easily evaluated. We can then find a solution iteratively by initializing the moments and then cycling through the variables updating each distribution in turn using (15).

## 3 CONTROLLING COMPLEXITY

The Relevance Vector framework provides a means for solving regression and classification problems in which we seek models which are highly sparse by selecting a subset from a larger pool of candidate kernel functions (one for each example in the training set). A key concept is the use of continuous hyperparameters to govern model complexity and thereby avoid the intractable problem of searching over an exponentially large discrete space of model structures. This approach, based on a hierarchical prior, was successfully used to find the optimal number of principal components in a Bayesian treatment of PCA [2].

A conventional way to remove superfluous parameters is to use a 'pruning' prior given by a Laplace distribution of the form

$$P(w) = \lambda \exp(-\lambda |w|). \qquad (16)$$

Unfortunately, such a choice of prior does not lead to a tractable variational treatment, since the corresponding variational solution given by (15) cannot be evaluated analytically.

Here we propose an alternative framework based on a hierarchical prior of the form

$$P(w|\alpha) = \mathcal{N}(w|0, \alpha^{-1}) \qquad (17)$$

as discussed previously, in which we use a hyperprior given by

$$P(\alpha) = \Gamma(\alpha|a,b) \equiv b^a \alpha^{a-1} e^{-b\alpha}/\Gamma(a) \qquad (18)$$

where $\Gamma(a)$ is the Gamma function. The distribution (18) has the useful properties

$$\langle \alpha \rangle = a/b, \qquad \langle \alpha^2 \rangle - \langle \alpha \rangle^2 = a/b^2. \qquad (19)$$

The marginal distribution of $w$ (a t-distribution) is then obtained by integrating over $\alpha$. A comparison of this marginal distribution, for $a = b = 1$, with the Laplace distribution (16) is shown in Figure 1.

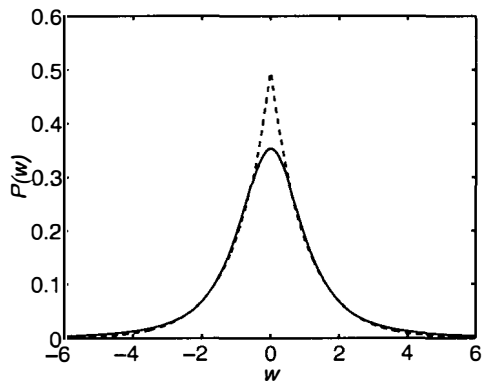

Figure 1: Comparison of the marginal distribution defined by the hierarchical model $P(w) = \int P(w|\alpha)P(\alpha)\,d\alpha$ (solid line), compared to the Laplace distribution (dotted line).

The key observation is that the variational framework can be rendered tractable by working not directly with the marginal distribution $P(w)$ but instead leaving the hierarchical conjugate form explicit and introducing a factorial representation given by $Q(w,\alpha) = Q(w)Q(\alpha)$. A further advantage of this approach is that it becomes possible to evaluate the lower bound $\mathcal{L}$ as a closed-form analytic expression. This is useful for monitoring the convergence of the iterative optimization and also for checking the accuracy of the software implementation (by verifying that none of the updates to the variational distributions lead to a decrease the value of $\mathcal{L}$). It can also be used to compare models (without resorting to a separate validation set) since it represents an approximation to the model evidence. We now exploit these ideas in the context of the Relevance Vector Machine.

## 4 RVM REGRESSION

Following the concepts developed in the previous section, we augment the standard relevance vector machine by the introduction of hyperpriors given by a separate distribution for each hyperparameter $\alpha_m$ of the form $P(\alpha_m) = \Gamma(\alpha_m|a,b)$. Similarly, we introduce a prior over the inverse noise variance $\tau$ given by $P(\tau) = \Gamma(\tau|c,d)$. We obtain broad hyperpriors by setting $a = b = c = d = 10^{-6}$. Together with the likelihood function (3) and the weight prior (4) we now have a complete probabilistic specification of the model. The probabilistic model can also be represented as a directed graph, as shown in Figure 2.

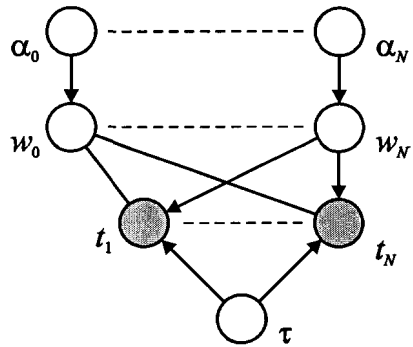

Figure 2: Directed acyclic graph representing the variational RVM as used for regression. The classification version is the same, with the omission of the $\tau$ node.

Next we consider a factorial approximation to the posterior distribution $P(\mathbf{w}, \boldsymbol{\alpha}, \tau | X, T)$ given by $Q(\mathbf{w}, \boldsymbol{\alpha}, \tau) = Q_\mathbf{w}(\mathbf{w})Q_{\boldsymbol{\alpha}}(\boldsymbol{\alpha})Q_\tau(\tau)$. Due to the conjugacy properties of the chosen distributions we can evaluate the general solution (15) analytically, giving

$$Q_\mathbf{w}(\mathbf{w}) = \mathcal{N}(\mathbf{w}|\boldsymbol{\mu}_\mathbf{w}, \boldsymbol{\Sigma}_\mathbf{w}) \qquad (20)$$

$$Q_\tau(\tau) = \Gamma(\tau|\widetilde{c}, \widetilde{d}) \qquad (21)$$

$$Q_{\boldsymbol{\alpha}}(\boldsymbol{\alpha}) = \prod_{m=0}^{N} \Gamma(\alpha_m|\widetilde{a}_m, \widetilde{b}_m) \qquad (22)$$

where

$$\boldsymbol{\Sigma}_\mathbf{w} = \left( \mathrm{diag}\langle \alpha_m \rangle + \langle \tau \rangle \sum_{n=1}^{N} \phi_n \phi_n^\mathrm{T} \right)^{-1} \qquad (23)$$

$$\boldsymbol{\mu}_\mathbf{w} = \langle \tau \rangle \boldsymbol{\Sigma}_\mathbf{w} \sum_{n=1}^{N} \phi_n t_n \qquad (24)$$

$$\widetilde{a}_m = a + 1/2 \qquad \widetilde{b}_m = b + \langle w_m^2 \rangle/2 \qquad (25)$$

$$\widetilde{c} = c + (N+1)/2 \qquad (26)$$

$$\widetilde{d} = d + \frac{1}{2} \sum_{n=1}^{N} t_n^2 - \langle \mathbf{w} \rangle^\mathrm{T} \sum_{n=1}^{N} \phi_n t_n$$

$$+ \frac{1}{2} \sum_{n=1}^{N} \phi_n^\mathrm{T} \langle \mathbf{w}\mathbf{w}^\mathrm{T} \rangle \phi_n. \qquad (27)$$

The required moments are easily evaluated using the

450 UNCERTAINTY IN ARTIFICIAL INTELLIGENCE PROCEEDINGS 2000

following results

$$\langle \mathbf{w} \rangle = \boldsymbol{\mu}_\mathbf{w} \tag{28}$$
$$\langle \mathbf{w}\mathbf{w}^T \rangle = \boldsymbol{\Sigma}_\mathbf{w} + \boldsymbol{\mu}_\mathbf{w}\boldsymbol{\mu}_\mathbf{w}^T \tag{29}$$
$$\langle \alpha_m \rangle = \widetilde{a}_m/\widetilde{b}_m \tag{30}$$
$$\langle \ln \alpha_m \rangle = \psi(\widetilde{a}_m) - \ln \widetilde{b}_m \tag{31}$$
$$\langle \tau \rangle = \widetilde{c}/\widetilde{d} \tag{32}$$
$$\langle \ln \tau \rangle = \psi(\widetilde{c}) - \ln \widetilde{d} \tag{33}$$

where the $\psi$ function is defined by

$$\psi(a) = \frac{d}{da}\ln \Gamma(a). \tag{34}$$

The full predictive distribution $P(t|\mathbf{x}, X, T)$ is given by

$$P(t|\mathbf{x}, X, T) = \iint P(t|\mathbf{x}, \mathbf{w}, \tau) P(\mathbf{w}, \tau|X, T) \, d\mathbf{w} \, d\tau. \tag{35}$$

In the variational framework we replace the true posterior $P(\mathbf{w}, \tau|X, T)$ by its variational approximation $Q_\mathbf{w}(\mathbf{w})Q_\tau(\tau)$. Integration over both $\mathbf{w}$ and $\tau$ is intractable. However, as the number of data points increases the distribution of $\tau$ becomes tightly concentrated around its mean value. To see this we note that the variance of $\tau$ is given, from (19), by $\langle \tau^2 \rangle - \langle \tau \rangle^2 = \widetilde{c}/\widetilde{d}^2 \sim O(1/N)$ for large $N$. Thus we can approximate the predictive distribution using

$$P(t|\mathbf{x}, X, T) = \int P(t|\mathbf{x}, \mathbf{w}, \langle \tau \rangle) Q_\mathbf{w}(\mathbf{w}) \, d\mathbf{w} \tag{36}$$

which is the convolution of two Gaussian distributions. Using (2) and (20) we then obtain

$$P(t|\mathbf{x}, X, T) = \mathcal{N}(t|\boldsymbol{\mu}_\mathbf{w}^T \boldsymbol{\phi}(\mathbf{x}), \sigma^2) \tag{37}$$

where the input-dependent variance is given by

$$\sigma^2(\mathbf{x}) = \frac{1}{\langle \tau \rangle} + \boldsymbol{\phi}(\mathbf{x})^T \boldsymbol{\Sigma}_\mathbf{w} \boldsymbol{\phi}(\mathbf{x}). \tag{38}$$

We can also evaluate the lower bound $\mathcal{L}$, given by (12), which in this case takes the form

$$\begin{aligned}\mathcal{L} &= \langle \ln P(T|X, \mathbf{w}, \tau) \rangle + \langle \ln P(\mathbf{w}|\boldsymbol{\alpha}) \rangle \\ &+ \langle \ln P(\boldsymbol{\alpha}) \rangle + \langle \ln P(\tau) \rangle - \langle \ln Q_\mathbf{w}(\mathbf{w}) \rangle \\ &- \langle \ln Q_{\boldsymbol{\alpha}}(\boldsymbol{\alpha}) \rangle - \langle \ln Q_\tau(\tau) \rangle \end{aligned} \tag{39}$$

in which

$$\begin{aligned}\langle \ln P(T|X, \mathbf{w}, \tau) \rangle &= \frac{N}{2}\langle \ln \tau \rangle - \frac{N}{2}\ln(2\pi) \\ &- \frac{1}{2}\langle \tau \rangle \left\{ \sum_{n=1}^N t_n^2 - 2\langle \mathbf{w} \rangle^T \sum_{n=1}^N \boldsymbol{\phi}_n t_n \right. \\ &\left. + \sum_{n=1}^N \boldsymbol{\phi}_n^T \langle \mathbf{w}\mathbf{w}^T \rangle \boldsymbol{\phi}_n \right\} \end{aligned} \tag{40}$$

$$\begin{aligned}\langle \ln P(\mathbf{w}|\boldsymbol{\alpha}) \rangle &= -\frac{N+1}{2}\ln(2\pi) - \frac{1}{2}\sum_{m=0}^N \langle \ln \alpha_m \rangle \\ &- \frac{1}{2}\sum_{m=0}^N \langle \alpha_m \rangle \langle w_m^2 \rangle \end{aligned} \tag{41}$$

$$\begin{aligned}\langle \ln P(\boldsymbol{\alpha}) \rangle &= (N+1)a \ln b + (a-1)\sum_{m=0}^N \langle \ln \alpha_m \rangle \\ &- b \sum_{m=0}^N \langle \alpha_m \rangle - (N+1)\ln \Gamma(a) \end{aligned} \tag{42}$$

$$\begin{aligned}\langle \ln P(\tau) \rangle &= c \ln d + (c-1)\langle \ln \tau \rangle \\ &- d\langle \tau \rangle - \ln \Gamma(c) \end{aligned} \tag{43}$$

$$\begin{aligned}-\langle \ln Q_\mathbf{w} \rangle &= (N+1)(1 + \ln(2\pi))/2 \\ &+ \ln |\boldsymbol{\Sigma}_\mathbf{w}|/2 \end{aligned} \tag{44}$$

$$\begin{aligned}-\langle \ln Q_{\boldsymbol{\alpha}} \rangle &= \sum_{m=0}^N \left\{ \widetilde{a}_m \ln \widetilde{b}_m + (\widetilde{a}_m - 1)\langle \ln \alpha_m \rangle \right. \\ &\left. - \widetilde{b}_m \langle \alpha_m \rangle - \ln \Gamma(\widetilde{a}_m) \right\} \end{aligned} \tag{45}$$

$$\begin{aligned}-\langle \ln Q_\tau \rangle &= \widetilde{c} \ln \widetilde{d} + (\widetilde{c} - 1)\langle \ln \tau \rangle \\ &- \widetilde{d}\langle \tau \rangle - \ln \Gamma(\widetilde{c}). \end{aligned} \tag{46}$$

Experimental results in which this framework is applied to synthetic and real data sets are given in Section 6.

## 5 RVM CLASSIFICATION

The classification case is somewhat more complex than the regression case since we no longer have a fully conjugate heirarchical structure. To see how to resolve this, consider again the log marginal probability of the target data, given the input data, which can be written

$$\ln P(T|X) = \ln \iint P(T|X, \mathbf{w})P(\mathbf{w}|\boldsymbol{\alpha})P(\boldsymbol{\alpha}) \, d\mathbf{w} \, d\boldsymbol{\alpha}. \tag{47}$$

As before we introduce a factorized variational posterior of the form $Q_\mathbf{w}(\mathbf{w})Q_{\boldsymbol{\alpha}}(\boldsymbol{\alpha})$, and obtain the following lower bound on the log marginal probability

$$\begin{aligned}\ln P(T|X) &\geq \iint Q_\mathbf{w}(\mathbf{w}) Q_{\boldsymbol{\alpha}}(\boldsymbol{\alpha}) \\ &\ln \left\{ \frac{P(T|X, \mathbf{w})P(\mathbf{w}|\boldsymbol{\alpha})P(\boldsymbol{\alpha})}{Q_\mathbf{w}(\mathbf{w})Q_{\boldsymbol{\alpha}}(\boldsymbol{\alpha})} \right\} d\mathbf{w} \, d\boldsymbol{\alpha}. \end{aligned} \tag{48}$$

Now, however, the right hand side of (48) is intractable. We therefore follow Jaakkola and Jordan [3] and introduce a further bound using the inequality

$$\sigma(y)^t[1 - \sigma(y)]^{1-t} = \sigma(z) \tag{49}$$
$$\geq \sigma(\xi)\exp\left(\frac{z - \xi}{2} - \lambda(\xi)(z^2 - \xi^2)\right) \tag{50}$$

where $z = (2t-1)y$ and $\lambda(\xi) = (1/4\xi)\tanh(\xi/2)$. Here $\xi$ is a variational parameter, such that equality is achieved for $\xi = z$. Thus we have

$$P(T|X, \mathbf{w}) \geq F(T, X, \mathbf{w}, \boldsymbol{\xi}) = \prod_{n=1}^{N} \sigma(\xi_n)$$
$$\exp\left(\frac{z_n - \xi_n}{2} - \lambda(\xi_n)(z_n^2 - \xi_n^2)\right) \quad (51)$$

where $z_n = (2t_n - 1)\mathbf{w}^T\boldsymbol{\phi}_n$. Substituting into (48), and noting that $P(T|X, \mathbf{w})/F(T, X, \mathbf{w}, \boldsymbol{\xi}) \geq 1$ implies $\ln P(T|X, \mathbf{w})/F(T, X, \mathbf{w}, \boldsymbol{\xi}) \geq 0$, we obtain a lower bound on the original lower bound, and hence we have

$$\ln P(T|X) \geq \mathcal{L} = \iint d\mathbf{w}\, d\boldsymbol{\alpha}\, Q_{\mathbf{w}}(\mathbf{w}) Q_{\boldsymbol{\alpha}}(\boldsymbol{\alpha})$$
$$\ln\left\{\frac{F(T, X, \mathbf{w})P(\mathbf{w}|\boldsymbol{\alpha})P(\boldsymbol{\alpha})}{Q_{\mathbf{w}}(\mathbf{w})Q_{\boldsymbol{\alpha}}(\boldsymbol{\alpha})}\right\}. \quad (52)$$

We now optimize the right hand side of (52) with respect to the functions $Q_{\mathbf{w}}(\mathbf{w})$ and $Q_{\boldsymbol{\alpha}}(\boldsymbol{\alpha})$ as well as with respect to the parameters $\boldsymbol{\xi} = \{\xi_n\}$. The variational optimization for $Q_{\mathbf{w}}(\mathbf{w})$ yields a normal distribution of the form

$$Q_{\mathbf{w}}(\mathbf{w}) = \mathcal{N}(\mathbf{w}|\mathbf{m}, \mathbf{S}) \quad (53)$$

$$\mathbf{S} = \left(\mathbf{A} + 2\sum_{n=1}^{N}\lambda(\xi_n)\boldsymbol{\phi}_n\boldsymbol{\phi}_n^T\right)^{-1} \quad (54)$$

$$\mathbf{m} = \frac{1}{2}\mathbf{S}\left(\sum_{n=1}^{N}(2t_n - 1)\boldsymbol{\phi}_n\right) \quad (55)$$

where $\mathbf{A} = \text{diag}\langle\alpha_m\rangle$. Similarly, variational optimization of $Q_{\boldsymbol{\alpha}}(\boldsymbol{\alpha})$ yields a product of Gamma distributions of the form

$$Q_{\boldsymbol{\alpha}}(\boldsymbol{\alpha}) = \prod_{m=0}^{N}\Gamma(\alpha_m|\widetilde{a}, \widetilde{b}_m) \quad (56)$$

$$\widetilde{a} = a + \frac{1}{2} \qquad \widetilde{b}_m = b + \frac{1}{2}\langle w_m^2\rangle. \quad (57)$$

Finally, maximizing (52) with respect to the variational parameters $\xi_n$ gives re-estimation equations of the form

$$\xi_n^2 = \boldsymbol{\phi}_n^T\langle\mathbf{w}\mathbf{w}^T\rangle\boldsymbol{\phi}_n. \quad (58)$$

We can also evaluate the lower bound given by the right hand side of (52)

$$\mathcal{L} = \langle\ln F\rangle + \langle\ln P(\mathbf{w}|\boldsymbol{\alpha})\rangle + \langle\ln P(\boldsymbol{\alpha})\rangle$$
$$- \langle\ln Q_{\mathbf{w}}(\mathbf{w})\rangle - \langle Q_{\boldsymbol{\alpha}}(\boldsymbol{\alpha})\rangle \quad (59)$$

where we have

$$\langle\ln F\rangle = \sum_{n=1}^{N}\left\{\ln\sigma(\xi_n) + \frac{1}{2}(2t_n - 1)\langle\mathbf{w}^T\rangle\boldsymbol{\phi}_n\right.$$
$$\left.-\frac{1}{2}\xi_n - \lambda(\xi_n)\left(\boldsymbol{\phi}_n^T\langle\mathbf{w}\mathbf{w}^T\rangle\boldsymbol{\phi}_n - \xi_n^2\right)\right\} \quad (60)$$

$$\langle\ln P(\mathbf{w}|\boldsymbol{\alpha})\rangle = -\frac{1}{2}\sum_{m=0}^{N}\langle\alpha_m\rangle\langle w_m^2\rangle$$
$$+\frac{1}{2}\sum_{m=0}^{N}\langle\ln\alpha_m\rangle - \frac{(N+1)}{2}\ln(2\pi) \quad (61)$$

$$\langle\ln P(\boldsymbol{\alpha})\rangle = \sum_{m=0}^{N}\left\{-b\widetilde{a}/\widetilde{b} + (a-1)\left(\psi(\widetilde{a}) - \ln\widetilde{b}\right)\right.$$
$$\left. + a\ln b - \ln\Gamma(a)\right\} \quad (62)$$

$$-\langle\ln Q_{\mathbf{w}}(\mathbf{w})\rangle = \frac{N+1}{2}(1 + \ln 2\pi) + \frac{1}{2}\ln|\mathbf{S}| \quad (63)$$

$$-\langle\ln Q_{\boldsymbol{\alpha}}(\boldsymbol{\alpha})\rangle = \sum_{m=0}^{N}\left\{-(\widetilde{a}_m - 1)\psi(\widetilde{a}_m)\right.$$
$$\left. - \ln\widetilde{b}_m + \widetilde{a}_m + \ln\Gamma(\widetilde{a}_m)\right\} \quad (64)$$

Predictions from the trained model for new inputs can be obtained by substituting the posterior mean weights into (8) to give the predictive distribution in the form

$$P(t|\mathbf{x}, \langle\mathbf{w}\rangle). \quad (65)$$

A more accurate estimate would take account of the weight uncertainty by marginalizing over the posterior distribution of the weights. Using the variational result $Q_{\mathbf{w}}(\mathbf{w})$ for the posterior distribution leads to convolution of a sigmoid with a Gaussian, which is intractable. From symmetry, however, such a marginalization does not change the location of the $p = 0.5$ decision surface. A useful approximation to the required integration has been given by MacKay [5].

## 6 EXPERIMENTAL RESULTS

### 6.1 REGRESSION

We illustrate the operation of the variational relevance vector machine (VRVM) for regression using first of all a synthetic data set based on the function $\text{sinc}(x) = (\sin x)/x$ for $x \in (-10, 10)$, with added noise. Figure 3 shows the result from a Gaussian kernel relevance vector regression model, and Figure 4 illustrates the mean hyperparameter values and weights associated with the model of Figure 3. Results from averaging over 25 such randomly generated data sets are shown in Table 1.

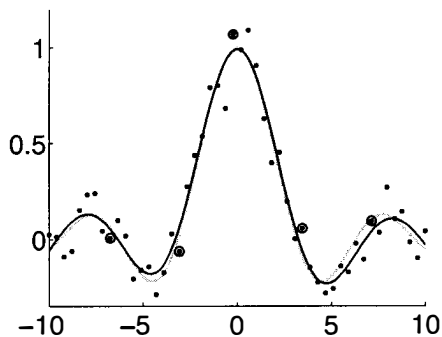

Figure 3: Example fit of a variational RVM to 50 data points generated from the 'sinc' function with added Gaussian noise of standard deviation 0.1. The sinc function and the mean interpolant are plotted in grey and black respectively, and the five relevance vectors (obtained by thresholding the mean weights at $10^{-3}$) are circled. The RMS deviation from the true function is 0.032, while a comparable SVM gave error of 0.038 using 36 support vectors. The VRVM also gives an estimate of the noise, which in this case had mean value 0.0945.

| Model | Error | # kernels | Noise estimate |
|-------|-------|-----------|----------------|
| SVM | 0.0519 | 28.0 | – |
| RVM | 0.0494 | 6.9 | 0.0943 |
| VRVM | 0.0494 | 7.4 | 0.0950 |

Table 1: RMS test error, number of utilised kernels and, for the relevance models, noise estimates averaged over 25 generations of the noisy sinc dataset. For all models, Gaussian kernels were used with the width parameter selected from a range of values using 5-fold cross-validation. For the SVM, the parameters $C$ (the trade-off parameter) and $\epsilon$ (controlling the insensitive region of the loss function) were chosen via a further 5-fold cross-validation.

| Model | Error | # kernels | Noise estimate |
|-------|-------|-----------|----------------|
| SVM | 10.29 | 235.2 | – |
| RVM | 10.17 | 41.1 | 2.49 |
| VRVM | 10.36 | 40.9 | 2.49 |

Table 2: Squared test error, number of utilised kernels and noise estimates averaged over 10 random partitions of the Boston housing dataset into training/test sets of size 481 and 25 respectively. A third order polynomial kernel was used.

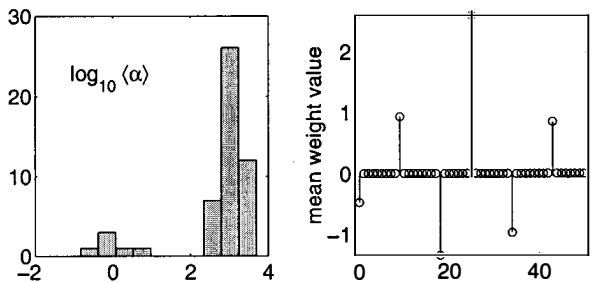

Figure 4: (Left) Histogram of the mean of the approximate $\alpha$ posterior. (Right) A plot of the 51 (unthresholded) mean weight values (the first weight is the bias, the next 50 correspond to the 50 data points, read left-to-right, in Figure 3). The dichotomy into 'relevant' and 'irrelevant' weights is clear.

As an example of a regression problem using real data, we show results in Table 2 for the popular Boston housing dataset.

### 6.2 CLASSIFICATION

We illustrate the operation of the VRVM for classification with some synthetic data in two dimensions taken from Ripley [7]. A randomly chosen subset of 100 training examples (of the original 250) was utilised to train an SVM, RVM and VRVM. Results from typical SVM and VRVM classifiers, using Gaussian kernels of width 0.5, are shown in Figures 5 and 6 respectively.

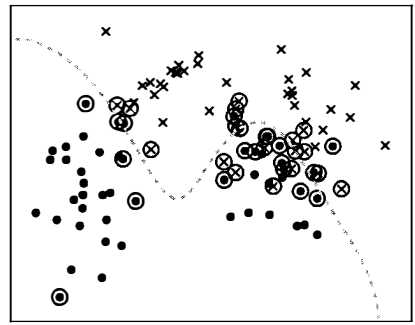

Figure 5: Support vector classifier of the Ripley dataset for which there are 38 kernel functions.

To assess the accuracy of the classifiers on this dataset, models with Gaussian kernels were used, with the width parameter of the Gaussian chosen by 5-fold cross-validation, and the SVM trade-off parameter $C$ was similarly estimated using a further 5-fold cross-validation. The results are given in Table 3.

The 'Pima Indians' diabetes dataset is a popular classification benchmark. Table 4 summarises results on Ripley's split of this dataset into 200 training and 332 test examples.

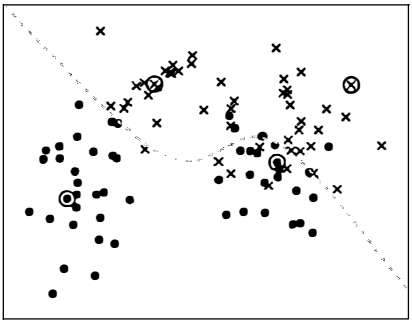

Figure 6: Variational relevance vector classifier of the Ripley dataset for which there are 4 kernel functions.

| Model | Error | # kernels |
|---|---|---|
| SVM | 10.6% | 38 |
| RVM | 9.3% | 4 |
| VRVM | 9.2% | 4 |

Table 3: Percentage misclassification rate and number of kernels used for classifiers on the Ripley synthetic data. The Bayes error rate for this data set is 8%.

| Model | Error | # kernels |
|---|---|---|
| SVM | 69 | 110 |
| RVM | 65 | 4 |
| VRVM | 65 | 4 |

Table 4: Number of misclassifications and number of kernels used for classifiers on the Pima Indians data.

## 7 DISCUSSION

In this paper we have developed a practical variational framework for the Bayesian treatment of Relevance Vector Machines.

The variational solution for the Relevance Vector Machine is computationally more expensive than the type-II maximum likelihood approach. However, the advantages of a fully Bayesian approach are expected to be most pronounced in situations where the size of the data set is limited, in which case the computational cost of the training phase is likely to be insignificant.